# Nepali Sign Language Characters Recognition: Dataset Development and Deep Learning Approaches

**Birat Poudel[1], Satyam Ghimire[2], Sijan Bhattarai[3], Saurav Bhandari[4], Suramya Sharma Dahal[5]**

[1]Department of Electronics and Computer Engineering, Thapathali Campus, IOE, TU, E-mail:poudel.birat25@gmail.com
[2]Department of Electronics and Computer Engineering, Thapathali Campus, IOE, TU, E-mail:satyamghimirestar@gmail.com
[3]Department of Electronics and Computer Engineering, Thapathali Campus, IOE, TU, E-mail:bhattaraisijan22@gmail.com
[4]Department of Electronics and Computer Engineering, Thapathali Campus, IOE, TU, E-mail:sauravb1234567@gmail.com
[5]Asssociate Professor, Dept of Electronics, Communication & Information Engineering, Kathmandu Engineering College, E-mail: suramya.sharma@kecktm.edu.np

*Abstract*— Sign languages serve as essential communication systems for individuals with hearing and speech impairments. However, digital linguistic dataset resources for underrepresented sign languages, such as Nepali Sign Language (NSL), remain scarce. This study introduces the first benchmark dataset for NSL, consisting of 36 gesture classes with 1,500 samples per class, designed to capture the structural and visual features of the language. To evaluate recognition performance, we fine-tuned MobileNetV2 and ResNet50 architectures on the dataset, achieving classification accuracies of 90.45% and 88.78%, respectively. These findings demonstrate the effectiveness of convolutional neural networks in sign recognition tasks, particularly within low-resource settings. To the best of our knowledge, this work represents the first systematic effort to construct a benchmark dataset and assess deep learning approaches for NSL recognition, highlighting the potential of transfer learning and fine-tuning for advancing research in underexplored sign languages.

**Introduction**

Communication for individuals with hearing and speech impairments remains a significant challenge in many parts of the world, especially where sign languages are under-resourced in terms of data, recognition, and technology. In Nepal, approximately 2.2% of the population are reported to have some form of disability according to the 2022 Census. Among these, about 7.85% are deaf, and an additional 7.87% have hearing impairment (hard-of-hearing), while 6.36% have speech impairment. [1] In total numbers, this means tens of thousands of individuals depend on sign language for daily communication, but there remain serious gaps in infrastructure, educational support, interpreters, and computational tools.

Nepali Sign Language (NSL) is the primary indigenous sign language used in Nepal. It has partially standardized forms, heavily influenced by the variety used in Kathmandu, along with contributions from other regions such as Pokhara. NSL is distinct from spoken Nepali; it is not simply a signed version of the oral language, but has its own grammar, lexicon, and sociolinguistic variation. [2] Historically, the first school for the deaf in Nepal was established in Kathmandu in 1966 by an ENT (ear-nose-throat) doctor, with an oralist approach (speech training) dominating the early decades. Later, organizations led by hearing and speech impaired people themselves (e.g. Kathmandu Association of the Deaf) pushed for inclusion of sign language, dictionaries, fingerspelling systems, and total communication approaches.

Despite this progress, NSL remains underrepresented in computational research. Unlike larger sign languages such as American or Indian Sign Language, there are no established benchmark datasets or systematic evaluations of recognition models for NSL. This absence limits the development of assistive technologies and hinders comparative research. Addressing this gap requires the creation of standardized datasets and the application of modern deep learning approaches to establish a foundation for future studies.

**Related Works**

A study from Bali, Indonesia of global sign-recognition methods concluded that data acquisition is foundational and that convolutional neural networks (CNNs) consistently yield the highest accuracy among evaluated techniques. [3]

Researchers from CHARUSAT, India trained sequential LSTM–GRU combinations on the IISL2020 dataset for Indian Sign Language and reported superior performance on common words. [4] A University of Algiers study applied YOLOv5 to Arabic Sign Language, finding higher detection accuracy than Faster R-CNN albeit with slower inference. [5] Dynamic sign-language work using a BLSTM–3D ResNet (3D residual ConvNet + bidirectional LSTM) achieved strong performance on long-term and visually similar gestures. [6] A 2017 ASL study trained a CNN on over 10,000 depth-annotated isolated and continuous signs (using supervised and unsupervised training variants) and reported ~93.3% accuracy with successful sentence-level recognition/translation. [7]

Pulchowk Engineering Campus (2015) implemented a contour-based hand-gesture pipeline (background subtraction → segmentation → feature extraction → classification) that yielded modest accuracy but guided later Nepali work. [8] Kantipur Engineering College developed an NSL detection system using red gloves (dataset referenced to the NSL dictionary) with two convolutional layers, a pooling layer and an ANN trained with mini-batch gradient descent, while noting glove-based limitations in generalization. [9] IOE (Pulchowk) students created an ASL recognition prototype with manually created datasets and a CNN using a truncated VGG-16 backbone to enable two-way text↔image communication. [10]

Taken together, these works show a clear progression from handcrafted contour- and glove-based systems toward deep-learning and multi-modal approaches (CNNs, RNNs, 3D CNNs, hybrid models), and they highlight the continued need for larger, standardized datasets and benchmarks for NSL and other under-resourced sign languages.

**Methodology**

A. Custom Dataset Creation

We created a custom dataset for Nepali Sign Language Character Recognition. The custom dataset contains images of 36 distinct Nepali Sign Language characters (classes 0-35), systematically collected and organized into two background variants:

1. Plain Background:
   a. Images with uniform, clean backgrounds for controlled learning conditions
   b. Each character has 1,000 images captured against a plain background
   c. Total: 36,000 images in plain background variant
2. Random Background:
   a. Images with varied, realistic backgrounds to enhance model robustness
   b. Each character has 500 images captured against a random background
   c. Total: 18,000 images in random background variant

This dual-background approach was intentionally designed to expose the network to diverse environmental conditions, simulating real-world deployment scenarios where sign language recognition must function across different lighting and background conditions. The combined dataset totals 54,000 images (36,000 plain + 18,000 random background), providing substantial training data with balanced class distribution across all 36 character classes.

The data preprocessing pipeline involved converting the raw image dataset into TensorFlow's TFRecord format, providing significant performance advantages for deep learning workflows like better I/O performance, storage efficiency, etc.

B. CNN Architecture, Parameters and Training Process

Convolutional Neural Network (CNN) is a type of deep neural network specifically designed for image processing tasks. When it comes to image classification, CNNs are the most popular and widely used architecture. This research utilizes two pre-trained CNN architectures for Nepali Sign Language character classification: MobileNetV2 and ResNet50.

Both MobileNetV2 and ResNet50 architecture has a base model pre-trained on ImageNet and custom classification head adapted for 36-class classification.

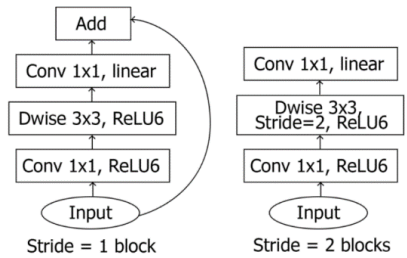

Fig. 1 MobileNetV2 Architecture

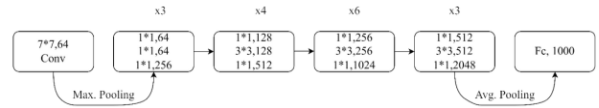

Fig. 2 ResNet50 Architecture

For multi-class classification, Sparse Categorical Cross Entropy was used in both architectures. Sparse categorical cross entropy is a loss function used in multi-class classification tasks where each example belongs to exactly one class. The loss function calculates:

$$LOSS = \sum_{i=1}^{output\ size} y_i . \log(\hat{y}_i) \qquad (1)$$

where,
$y_i$ = corresponding target value,
$\hat{y}_i$ = $i^{th}$ scalar value in the model output,
output size = number of scalar values in the model output

To maximize accuracy, transfer learning and fine-tuning were implemented using a progressive three-phase training strategy. Transfer learning utilizes models pre-trained on ImageNet for the sign language classification task. Fine-tuning further improves performance by gradually unfreezing layers and training with progressively reduced learning rates, allowing the model to adapt to sign language features while retaining learned representations. This progressive two-phase approach ensures optimal knowledge transfer from ImageNet features to Nepali Sign Language characteristics, maximizing classification accuracy while preventing overfitting through careful learning rate scheduling and gradual layer unfreezing.

The model training was conducted in two progressive phases using pre-trained architectures (MobileNetV2 and ResNet50) as the base networks. The key steps are summarized as follows:

1. Model Initialization: Pre-trained models were loaded and initialized to leverage transfer learning.
2. Hyperparameter Definition: Core hyperparameters, including the number of epochs, learning rate, and batch size, were defined prior to training.
3. Phase 1 - Frozen Base Model Training: The convolutional base was kept frozen, and only the newly added classifier layers were trained with a learning rate of 0.005 for 10 epochs.
4. Phase 2 - Partial Fine-Tuning: Selected deeper layers of the base model were unfrozen for fine-tuning, with training performed at a reduced learning rate of 0.00001 for 10 epochs.
5. Performance Monitoring: Accuracy and loss metrics were continuously monitored across epochs.
6. Final Evaluation: The trained models were evaluated using standard classification metrics, including precision, recall, and F1-score, to comprehensively assess their performance.

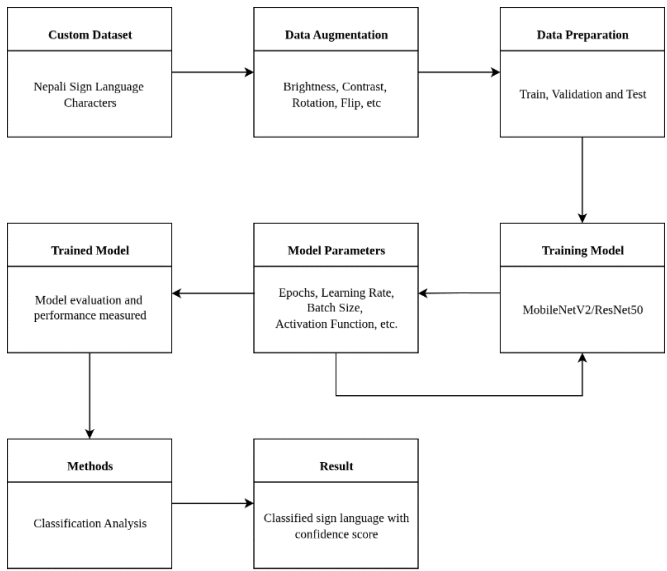

Fig. 3 Training Block Diagram

C. Evaluation Metrics

Model performance was assessed using a confusion matrix along with standard classification metrics: precision, recall, and F1-score. The confusion matrix provides a comprehensive summary of predicted versus actual classifications, where each row corresponds to the actual class and each column corresponds to the predicted class.

Precision measures the proportion of correctly predicted positive instances among all predicted positives.

$$Precision = \frac{TP}{TP + FP} \quad (2)$$

Recall (or sensitivity) quantifies the proportion of correctly identified positive instances among all actual positives.

$$Recall = \frac{TP}{FN + TP} \quad (3)$$

F1-Score represents the harmonic mean of precision and recall, balancing both metrics to provide a single measure of classification performance.

$$F1 - score = \frac{2*precision*recall}{precision+recall} \quad (4)$$

These metrics were used collectively to evaluate the robustness and accuracy of the proposed models in recognizing Nepali Sign Language characters.

D. System Methodology

The proposed system architecture implements a complete pipeline for Nepali Sign Language character recognition, integrating both model inference and user interaction through a web-based interface.

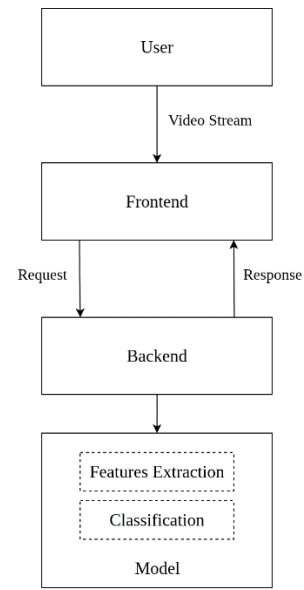

Fig. 4 System Block Diagram

1. User Input: Users upload sign language character images via a web application interface.
2. Preprocessing: The server applies preprocessing steps such as resizing and normalization to prepare the images for model input.
3. Feature Extraction: Pre-trained convolutional neural networks (MobileNetV2 and ResNet50) are employed to extract high-level image features.
4. Classification: Extracted features are passed through a custom dense classification head to predict the target class (characters 0–35).

5. Prediction Output: The system generates real-time predictions along with confidence scores, which are displayed on the user interface.

The backend is implemented using FastAPI, which efficiently handles model loading, inference requests, and communication with the user interface. This design ensures robust and accurate classification while maintaining computational efficiency through optimized data pipelines and transfer learning techniques.

**Results and analysis**

We trained a diverse range of models, varying parameters such as batch size, learning rate, and the number of epochs. We also experimented with different architectural configurations by adding and adjusting various layers. Through this iterative process, we successfully developed a high-accuracy model. Across all experiments, we consistently used the Adam optimizer and the sparse categorical cross-entropy loss function.

Best Nepali Sign Language Characters Recognition
When it came to Nepali Sign Language characters recognition, MobileNetV2 out performed ResNet50. The same parameters and layers were used for both models and their performance was compared.

TABLE 1

MOBILENETV2 AND RESNET50 ACCURACY AND LOSS FOR NEPALI SIGN LANGUAGE CHARACTERS CLASSIFICATION

| Model | Training Loss | Training Accuracy | Validation Loss | Validation Accuracy |
|---|---|---|---|---|
| MobileNetV2 | 0.3543 | 90.45% | 0.0373 | 98.99% |
| ResNet50 | 0.4252 | 88.78% | 0.1162 | 96.70% |

TABLE 2

MODEL HYPERPARAMETERS FOR NEPALI SIGN LANGUAGE CHARACTERS CLASSIFICATION

| Model Hyperparameter | Transfer Learning | Fine-tunning |
|---|---|---|
| Learning Rate | 0.005 | 0.00001 |
| Optimizer | Adam | Adam |
| Batch Size | 32 | 32 |
| Epochs | 10 | 10 |
| Loss Function | Sparse Categorical Cross Entropy | Sparse Categorical Cross Entropy |

TABLE 3

MODEL LAYERS FOR NEPALI SIGN LANGUAGE CHARACTERS CLASSIFICATION

| Layer Number | Layer Type | Layer Parameters |
|---|---|---|
| 1 | Global Average Pooling 2D | |
| 2 | Dense | 256 |
| 3 | Activation | ReLu |
| 4 | Dropout | 0.5 |
| 5 | Dense | 128 |
| 6 | Activation | ReLu |
| 7 | Dropout | 0.3 |
| 9 | Dense | 36 |

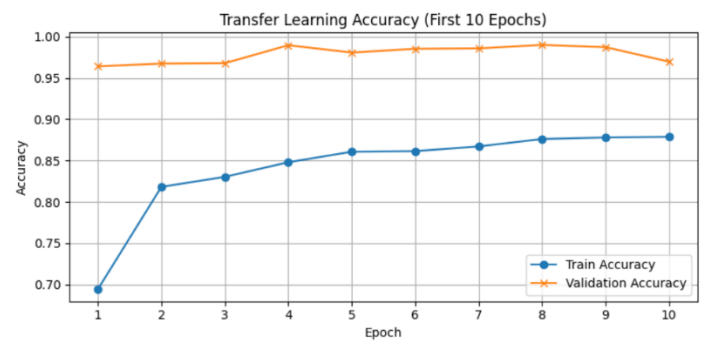

Fig. 5 MobileNetV2 Transfer Learning Accuracy (First 10 Epochs) for NSL Characters Classification

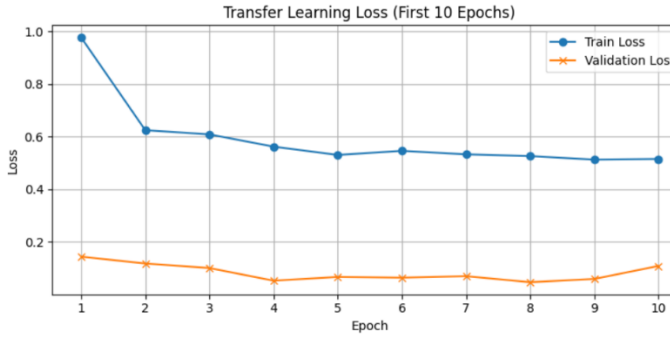

Fig. 6 MobileNetV2 Transfer Learning Loss (First 10 Epochs) for NSL Characters Classification

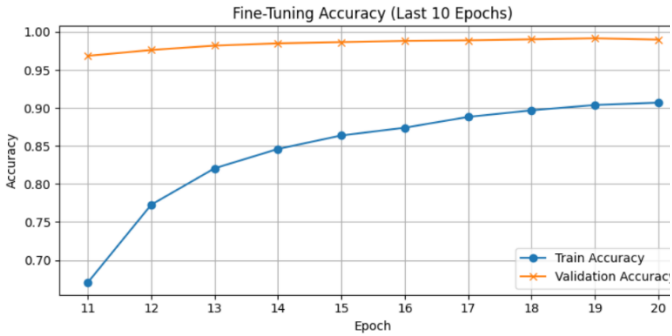

Fig. 7 MobileNetV2 Fine-Tuning Accuracy (Last 10 Epochs) for NSL Characters Classification

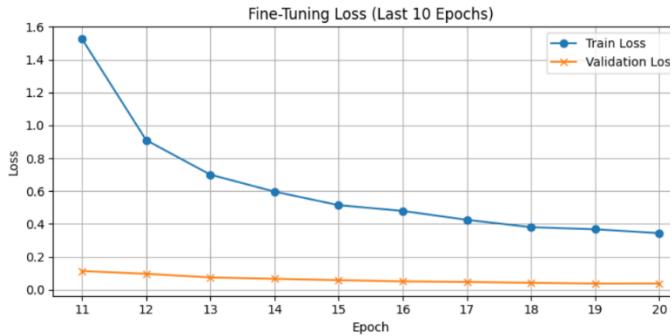

Fig. 8 MobileNetV2 Fine-Tuning Loss (Last 10 Epochs) for NSL Characters Classification

A sharp drop in accuracy or a sudden spike in loss is commonly observed at the onset of fine-tuning. This behavior occurs because, once the previously frozen layers are unfrozen, the model transitions from relying solely on pretrained weights to adapting those weights for the specific dataset. During this phase, the optimizer must readjust a large number of parameters, which often leads to temporary instability in performance. However, with continued training, the model typically stabilizes, allowing it to recover and often achieve higher accuracy than in the frozen-base stage.

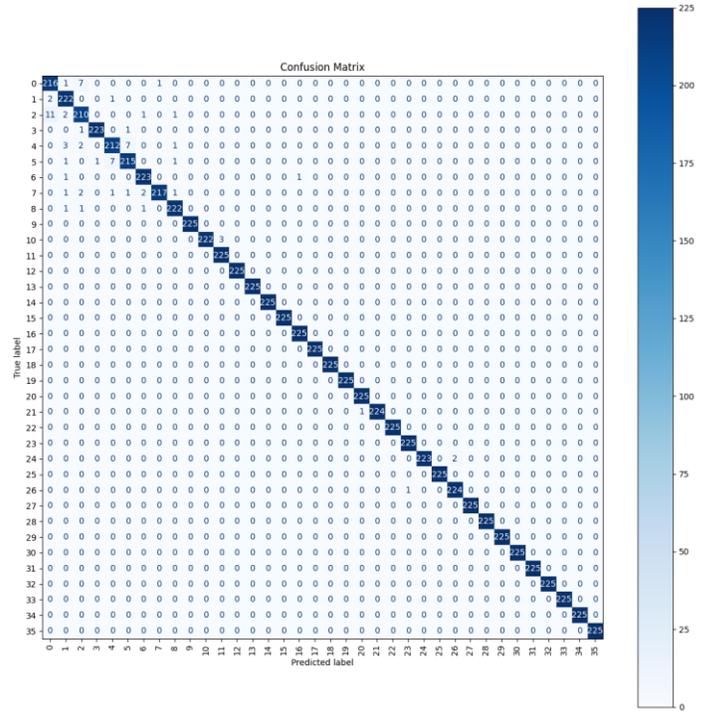

Fig. 9 Confusion Matrix for MobileNetV2 NSL Characters Classification

TABLE 4

PRECISION, RECALL AND F1 SCORE COMPARISON FOR MOBILENETV2 NSL CHARACTERS CLASSIFICATION

| Precision | Recall | F1-Score | Support |
|---|---|---|---|
| Accuracy | | 0.99 | 8100 |
| Macro Average | 0.99 | 0.99 | 8100 |
| Weighted Average | 0.99 | 0.99 | 8100 |

### A. Comparison between MobileNetV2 and ResNet50

The results indicated that, in general, MobileNetV2 achieved higher performance than ResNet50 for Nepali Sign Language (NSL) characters recognition. While ResNet50 has often outperformed other models on large-scale datasets such as ImageNet, our experiments showed that MobileNetV2 was more effective in this low-resource setting, achieving a classification accuracy of 90.45% compared to 88.78% for ResNet50. This outcome is consistent with findings from other studies where lightweight architectures can surpass deeper models when the dataset size and visual complexity are moderate. In our case, the dataset comprised 37,800 training images, 8,100 validation images, and 8,100 test images across 36 gesture classes.

There are several possible reasons for this performance difference. MobileNetV2 is designed as a lightweight architecture with significantly fewer parameters than ResNet50, reducing the risk of overfitting when trained on medium-scale datasets. Its use of depthwise separable convolutions allows it to efficiently capture localized spatial and structural features, which are crucial for distinguishing hand gestures in NSL. Since these gestures primarily involve variations in hand shape and orientation, MobileNetV2 is well-suited to capture the relevant discriminative patterns without unnecessary complexity.

In contrast, ResNet50, with its deeper architecture and larger parameter count, may have extracted redundant or non-essential features that do not contribute to the classification of relatively simple gesture images. The additional layers can be advantageous for highly complex datasets, but in this case, they may have led to reduced generalization compared to MobileNetV2. Consequently, MobileNetV2 not only trained more efficiently but also generalized better to unseen test data, as reflected in its superior accuracy.

Thus, although deeper networks like ResNet50 often dominate benchmarks on highly complex visual datasets, our findings demonstrate that for NSL recognition, MobileNetV2 offers a more effective balance between architectural efficiency, dataset size, and task complexity.

### B. Model Prediction

A continuous video stream of hand gestures is provided as input to the system. From this stream, frames are sampled at fixed intervals (e.g., every 3rd frame) to reduce redundancy and computational overhead. Each sampled frame is preprocessed and classified into one of the predefined Nepali Sign Language gesture classes using the trained model. To address changes in gestures and prevent stale predictions, a sliding window of consecutive frame predictions is maintained, and the final recognized gesture is determined using majority voting within this window. This approach ensures robust and accurate real-time recognition of gestures from the video stream.

## Conclusion

In this study, we presented the first benchmark dataset for Nepali Sign Language (NSL) comprising 36 gesture classes with 1,500 samples each, capturing the structural and visual nuances of the language. We evaluated the performance of deep learning models, specifically MobileNetV2 and ResNet50, using transfer learning and fine-tuning, achieving classification accuracies of 90.45% and 88.78%, respectively. The results demonstrate that convolutional neural networks are effective for real-time gesture recognition in low-resource settings. Additionally, our video-based frame extraction and sliding-window approach provides robust and stable recognition, addressing challenges related to gesture transitions and transient frames. Overall, this work lays the foundation for developing NSL recognition systems and contributes a valuable dataset for future research.

## Future Enhancements

Future research can pursue multiple avenues to strengthen the system. Increasing the variety of gesture classes and the number of samples would improve the model's ability to generalize across different users and contexts. Incorporating more advanced neural network architectures can more effectively capture the details and dynamics of gestures. Efforts to streamline the system for deployment on mobile and edge devices would enable faster processing and broader accessibility. Furthermore, combining hand gestures with additional modalities, such as facial expressions or body posture, could enhance the system's overall recognition accuracy and reliability.